# Nugget Proposal Networks for Chinese Event Detection

**Hongyu Lin**[1,2], **Yaojie Lu**[1,2], **Xianpei Han**[1], **Le Sun**[1]
[1]State Key Laboratory of Computer Science
Institute of Software, Chinese Academy of Sciences, Beijing, China
[2]University of Chinese Academy of Sciences, Beijing, China
{hongyu2016,yaojie2017,xianpei,sunle}@iscas.ac.cn

## Abstract

Neural network based models commonly regard event detection as a word-wise classification task, which suffer from the mismatch problem between words and event triggers, especially in languages without natural word delimiters such as Chinese. In this paper, we propose Nugget Proposal Networks (NPNs), which can solve the word-trigger mismatch problem by directly proposing entire trigger nuggets centered at each character regardless of word boundaries. Specifically, NPNs perform event detection in a character-wise paradigm, where a hybrid representation for each character is first learned to capture both structural and semantic information from both characters and words. Then based on learned representations, trigger nuggets are proposed and categorized by exploiting character compositional structures of Chinese event triggers. Experiments on both ACE2005 and TAC KBP 2017 datasets show that NPNs significantly outperform the state-of-the-art methods.

## 1 Introduction

Automatic event extraction is a fundamental task of information extraction. Event detection, which aims to identify event triggers of specific types, is a key step of event extraction. For example, from the sentence "*Henry was injured, and then passed away soon*", an event detection system should detect an "*Injure*" event triggered by "*injured*", and a "*Die*" event triggered by "*passed away*".

Recently, neural network methods, which transform event detection into a word-wise classification paradigm, have achieved significant progress in event detection (Nguyen and Grishman, 2015;

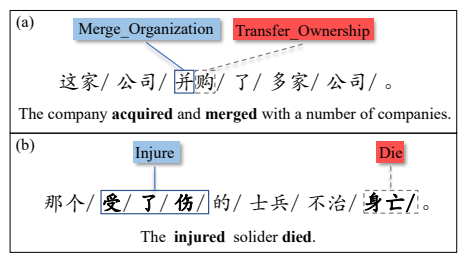

Figure 1: Examples of word-trigger mismatch. Slashes in the figure indicate word boundaries.

Chen et al., 2015b; Ghaeini et al., 2016). For instance, a model will detect events in sentence "*Henry was injured*" by successively classifying its three words into *NIL*, *NIL* and *Injure*. By automatically extracting features from raw texts, these methods rely little on prior knowledge and achieved promising results.

Unfortunately, word-wise event detection models suffer from the word-trigger mismatch problem, because a number of triggers do not exactly match with a word. Specifically, a trigger can be part of a word or cross multiple words, which is impossible to detect using word-wise models. This problem is more severe in languages without natural word delimiters such as Chinese. Figure 1 (a) shows several examples of part-of-word triggers, where two characters in one word "并购"(acquire and merge) trigger two different events: a "*Merge_Org*" event triggered by "并"(merge) and a "*Transfer_Ownership*" event triggered by "购" (acquire). Figure 1 (b) shows a multi-word trigger, where three words "受"(is), "了" and "伤"(injured) trigger an *Injure* event together. Table 1 shows the statistics of different types of word-trigger match on two standard datasets. We can see that word-trigger mismatch is crucial for Chinese event detection since nearly 25% of triggers in RichERE and 15% of them in ACE2005 dataset don't exactly match with a word.

To resolve the word-trigger mismatch problem,

| Match Type | Rich ERE | ACE2005 |
|---|---|---|
| Exact Match | 75.52% | 85.39% |
| Part of Word | 19.55% | 11.67% |
| Cross words | 4.93% | 2.94% |

Table 1: Percentages of different types of matches between words and triggers.

this paper proposes *Nugget Proposal Networks (NPNs)*, which identify triggers by modeling character compositional structures of trigger nuggets regardless of word boundaries. Given a sentence, NPNs regard characters as basic detecting units and are able to 1) directly propose the entire potential trigger nugget at each character by exploiting inner compositional structure of triggers; 2) effectively categorize proposed triggers by learning semantic representation from both characters and words. For example, at character "伤"(injured) in Figure 1 (b), NPNs are not only capable to detect it is part of an *Injure* event trigger, but also can propose the entire trigger nugget "受了伤"(is injured). The main idea behind NPNs is that most Chinese triggers have regular character compositional structure (Li et al., 2012). Concretely, most of Chinese event triggers have one central character which can indicate its event type, e.g. "杀"(kill) in "枪杀"(kill by shooting). Furthermore, characters are composed into a trigger based on regular compositional structures, e.g. "manner + verb" for "枪杀"(kill by shooting), "砍杀"(hack to death), as well as "verb + auxiliary + noun" for "受了伤"(is injured) and "挨了打"(beaten).

Figure 2 shows the architecture of NPNs. Given a character in sentence, a hybrid representation learning module is first used to learn its semantic representation from both characters and words in the sentence. This hybrid representation is then fed into two modules: one is *trigger nugget generator*, which proposes the entire potential trigger nugget by exploiting inner character compositional structure. Once a trigger is proposed, an *event type classifier* is applied to determine its event type. Compared with previous methods, NPNs mainly have following advantages:

1) **By directly proposing the entire trigger nugget centered at a character, trigger nugget generator can effectively resolve the word-trigger mismatch problem.** First, using characters as basic units, NPNs will not suffer from the word-trigger mismatch problem of word-wise methods. Furthermore, by modeling and exploiting character compositional structure of triggers, our model is more error-tolerant to character-wise classification errors than traditional character-based models, as shown in Section 4.4.

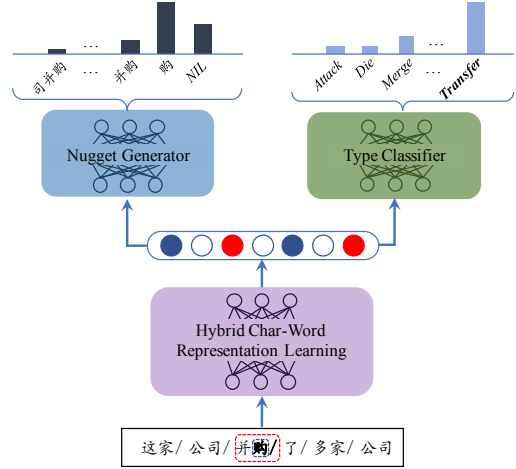

Figure 2: The overall architecture of Nugget Proposal Networks. The concerning character is "购".

2) **By summarizing information from both characters and words, our hybrid representation can effectively capture information for both inner character composition and accurate event categorization.** For example, the inner compositional structure of trigger "枪杀"(kill by shooting) can be learned from the character-level sequence. Besides, characters are often ambiguous, therefore the accurate representations must take their word context into consideration. For example, the representation "杀"(kill) in "枪杀"(kill by shooting) should be different from its representation in "杀青"(completed).

We conducted experiments on both the ACE2005 and the TAC KBP 2017 Event Nugget Detection datasets. Experiment results show that NPNs can effectively solve the word-mismatch problem, and therefore significantly outperform previous state-of-the-art methods[1].

## 2 Hybrid Representation Learning

Given a sentence, NPNs will first learn a representation for each character, then the representation is fed into downstream modules. We observe that both characters and words contain rich information for Chinese event detection: characters reveals the inner compositional structure of event triggers (Li et al., 2012), while words can provide more accurate and less ambiguous semantics than characters (Chen et al., 2015a). For example, character-level information can tell us that "枪杀"(kill by shooting) is a trigger constructed of regular pattern "manner + verb". While word-level sequences can provide more explicit information when we distinguish the semantics of "杀"(kill) in this context with that character in other words like "杀青"(completed).

---

[1] Our source code, including all hyper-parameter settings and pre-trained word embeddings, is openly available at github.com/sanmusunrise/NPNs.

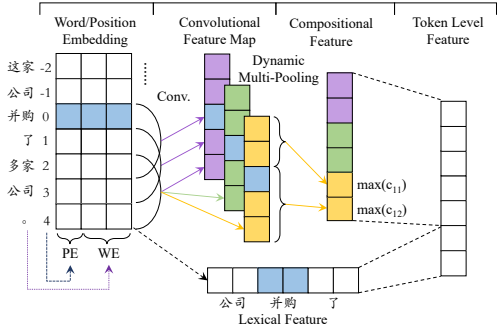

Figure 3: Token-level feature extractor, where PE is relative positional embeddings and WE is word embeddings. The concerning token is "并购".

Therefore, we propose to learn a hybrid representation which can summarize information from both characters and words. Specifically, we first learn two separate character-level and word-level representations using token-level neural networks. Then we design three kinds of hybrid paradigms to obtain the hybrid representation.

### 2.1 Token-level Representation Learning

Two token-level neural networks are used to extract features from characters and words respectively. The network architecture is similar to DMCNN (Chen et al., 2015b). Figure 3 shows a word-level example. Given $n$ tokens $t_1, t_2, ..., t_n$ in the sentence and the concerning token $t_c$, let $\mathbf{x}_i$ be the concatenation of the word embedding of $t_i$ and the embedding of $t_i$'s relative position to $t_c$, a convolutional layer with window size as $h$ is introduced to capture compositional semantics:

$$r_{ij} = \tanh(\mathbf{w}_i \cdot \mathbf{x}_{j:j+h-1} + b_i) \quad (1)$$

Here $\mathbf{x}_{i:i+j}$ refers to the concatenation of embeddings from $\mathbf{x}_i$ to $\mathbf{x}_{i+j}$, $\mathbf{w}_i$ is the i-th filter of the convolutional layer, $b_i \in R$ is a bias term. Then a dynamic multi-pooling layer is applied to preserve important signals of different parts of the sentence:

$$r_i^{left} = \max_{j<c} r_{ij}, \quad r_i^{right} = \max_{j \geq c} r_{ij} \quad (2)$$

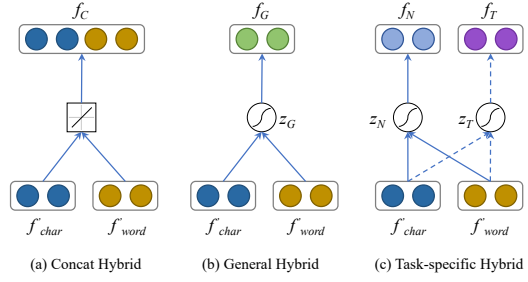

Figure 4: Three hybrid representation learning methods.

After that we concatenate $r_i^{left}$ and $r_i^{right}$ from all feature maps, as well as the embeddings of tokens nearing to $t_c$ to obtain the word-level representation $\mathbf{f_{word}}$ of $t_c$. Using the same procedure to character sequences, we can obtain the character-level representation $\mathbf{f_{char}}$.

### 2.2 Hybrid Representation Learning

So far we have both character-level feature representation $\mathbf{f_{char}}$ and word-level feature representation $\mathbf{f_{word}}$. This section describes how we mix them up to obtain a hybrid representation. Before this, we first project $\mathbf{f_{char}}$ and $\mathbf{f_{word}}$ respectively into the same vector space using two dense layers, and we represent the projected $d'$-dimensional vectors as $\mathbf{f'_{char}}$ and $\mathbf{f'_{word}}$. Then we design three different paradigms to mix them up: Concat Hybrid, General Hybrid and Task-specific Hybrid, as illustrated in Figure 4.

**Concat Hybrid** is the most simple method, which simply concatenates character-level and word-level representations:

$$\mathbf{f_C} = \mathbf{f'_{char}} \oplus \mathbf{f'_{word}} \quad (3)$$

This simple approach doesn't introduce any additional parameter, but we find it very effective in our experiments.

**General Hybrid** aims to learn a shared hybrid representation for both trigger nugget proposal and event type classification. Specifically, we design a gated structure to model the information flow from $\mathbf{f'_{char}}$ and $\mathbf{f'_{word}}$ to the general hybrid feature representation $\mathbf{f_G}$:

$$\mathbf{z_G} = s(\mathbf{W_{GH}} \mathbf{f'_{char}} + \mathbf{U_{GH}} \mathbf{f'_{word}} + \mathbf{b_{GH}}) \quad (4)$$

$$\mathbf{f_G} = \mathbf{z_G} \mathbf{f'_{char}} + (\mathbf{1} - \mathbf{z_G}) \mathbf{f'_{word}} \quad (5)$$

Here $s$ is the sigmoid function, $\mathbf{W_{GH}} \in R^{d' \times d'}$ and $\mathbf{U_{GH}} \in R^{d' \times d'}$ are weight matrix, and

$\mathbf{b_{GH}} \in R^{d'}$ is the bias term. $\mathbf{z_G}$ is a $d'$-dimensional vector whose values represent the contribution of $\mathbf{f'_{char}}$ and $\mathbf{f'_{word}}$ to the final hybrid representation, which models the importance of individual features in the given contexts.

As two downstream modules of NPNs have individual functions, they might hold different requirements to the input features. Intuitively, trigger nugget generator depends more on fine-grained character-level features. In contrast, word-level features might play more important roles in the event type classifier since it is enriched with more explicit semantics. As a result, a unified representation may be insufficient and it is better to learn task-specific hybrid representations.

**Task-specific Hybrid** is proposed to tackle this problem, where two gates are introduced for two modules respectively. Formally, we learn one representation for the trigger nugget generator and one for event type classifier as:

$$\mathbf{z_N} = s(\mathbf{W_N f'_{char}} + \mathbf{U_N f'_{word}} + \mathbf{b_N}) \quad (6)$$

$$\mathbf{z_T} = s(\mathbf{W_T f'_{char}} + \mathbf{U_T f'_{word}} + \mathbf{b_T}) \quad (7)$$

$$\mathbf{f_N} = \mathbf{z_N f'_{char}} + (\mathbf{1} - \mathbf{z_N})\mathbf{f'_{word}} \quad (8)$$

$$\mathbf{f_T} = \mathbf{z_T f'_{char}} + (\mathbf{1} - \mathbf{z_T})\mathbf{f'_{word}} \quad (9)$$

Here $\mathbf{f_N}$ and $\mathbf{f_T}$ are hybrid features for the trigger nugget generator and the event type classifier respectively and the meanings of other parameters are similar to the ones in Equation (4) and (5).

## 3 Nugget Proposal Networks

Given the hybrid representation of a character in a sentence, the goal of NPNs is to propose the potential trigger nugget, as well as to identify its corresponding event type at each character. For example in Figure 5, centered at the character "伤"(injured), NPNs need to propose "受了伤"(is injured) as the entire trigger nugget and identify its event type as "*Injure*". For this, NPNs are equipped with two modules: one is called *trigger nugget generator*, which is used to propose the potential trigger nugget containing the concerning character by exploiting character compositional structures of triggers. Another module, named as *event type classifier*, is used to determine the specific type of this event once a trigger nugget is detected.

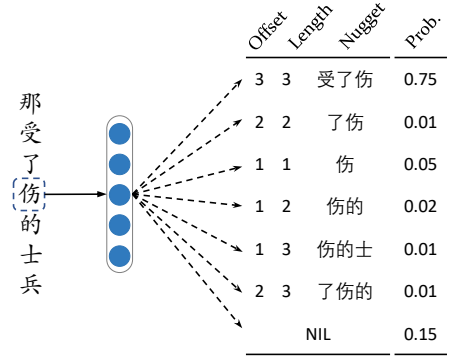

Figure 5: Our trigger nugget generator. For each character, there are 7 candidate nuggets including "NIL" if the maximum length of nuggets is 3.

### 3.1 Trigger Nugget Generator

Chinese event triggers have regular inner compositional structures, e.g. "受了伤"(is injured) and "挨了打"(is beaten) have the same "verb + auxiliary + noun" structure, and "枪杀"(kill by shooting) and "射杀"(kill by shooting) share the same "manner + verb" pattern. If a model is able to learn this compositional structure regularity, it can effectively detect trigger nuggets at characters. Recent advances have presented that convolutional neural networks are effective at capturing and predicting the region information in object detection (Ren et al., 2015) and semantic segmentation (He et al., 2017), which reveals the strong ability of CNNs to learning spatial and positional information. Inspired by this, we propose a neural network based trigger nugget generator, which is expected to not only be able to predict whether a character belongs to a trigger nugget, but also can point out the entire trigger nugget.

Figure 5 is an illustration of our trigger nugget generator. Hybrid representation $\mathbf{f_N}$ for concerning character is first learned as described in Section 2, which is then fed into a fully-connected layer to compute the scores for different possible trigger nuggets containing that character:

$$\mathbf{O^G} = \mathbf{W_G f_N} + \mathbf{b_G} \quad (10)$$

where $\mathbf{O^G} \in R^{d^N}$ and $d^N$ is the amount of candidate nuggets plus one *"NIL"* label indicating this character doesn't belong to an trigger. Given the maximum length $L$ of trigger nuggets, there are $\frac{L^2+L}{2}$ possible nuggets containing a specific character, as we shown in Figure 5. In both ACE and Rich ERE corpus, more than 98.5% triggers contain no more than 3 characters, so for a specific character we consider 6 candidate nuggets and thus $d^N = 7$. We expect NPNs to give a high score to a nugget if it follows a regular compositional structure of triggers. For example in Figure 5, "受了伤"(is injured) follows the compositional pattern of "verb + auxiliary + noun", therefore a high score is given to the category where "伤" is at the 3$^{rd}$ place of a nugget with a length of 3. By contrast "了伤" does not match a regular pattern, then the score for "伤" at the 2$^{nd}$ place of a nugget with a length of 2 will be low in this context.

After obtaining the scores for each nugget, a softmax layer is applied to normalize the scores:

$$P(y_i^G|x;\theta) = \frac{e^{O_i^G}}{\sum_{j=1}^{d^N} e^{O_j^G}} \quad (11)$$

where $O_i^G$ is the i-the element in $\mathbf{O^G}$ and $\theta$ is the model parameters.

### 3.2 Event Type Classifier

The event type classifier aims to identify whether the given character in the given context will exhibit an event type. Once we detect an event trigger nugget at one character, the hybrid feature $\mathbf{f_T}$ extracted previously is then feed into a neural network classifier, which further determines the specific type of this trigger. Following previous work (Chen and Ng, 2012), our event type classifier directly classifies nuggets into event subtypes, while ignores the hierarchy between event types.

Formally, given the hybrid feature vector $\mathbf{f_T}$ of input $x$, a fully-connected layer is applied to compute its scores assigned to each event subtype:

$$\mathbf{O^C} = \mathbf{W_C}\mathbf{f_T} + \mathbf{b_C} \quad (12)$$

where $\mathbf{O^C} \in R^{d^T}$ and $d^T$ is the number of event subtypes. Then similar to the trigger nugget generator, a softmax layer is introduced:

$$P(y_i^C|x;\theta) = \frac{e^{O_i^C}}{\sum_{j=1}^{d^T} e^{O_j^C}} \quad (13)$$

where $O_i^C$ is the i-th element in $\mathbf{O^C}$, representing the score for i-th subtype.

### 3.3 Dealing with Conflicts between Proposed Nuggets

While NPNs directly propose nugget at each character, there might exists conflicts between proposed nuggets at different characters. Generally speaking, there are two types of conflicts: (i) NIL/trigger conflict, which means NPNs propose a trigger nugget at one character, but classify other character in that nugget into "NIL" (e.g., proposing nugget "受了伤"(is injured) at "受" and output "NIL" at "了"); (ii) overlapped conflict, i.e., proposing two overlapped nuggets (e.g., proposing nugget "受了伤"(is injured) at "受" and nugget "伤" at "伤"). But we find that overlapped conflict is very rare because NPNs is very effective in capturing positional knowledge and the main challenge of event detection is to distinguish triggers from non-triggers.

Therefore in this paper, we employ a redundant prediction strategy by simply adding all proposed nuggets into results and ignoring "NIL" predictions. For example, if NPNs successively propose "受了伤"(is injured), "NIL", " 伤" from "受了伤", then we will ignore the "NIL" and add both two other nuggets into result. We found such a redundant prediction paradigm is an advantage of our model. Compared with conventional character-based models, even NPNs mistakenly classified character "了" into "NIL", we can still accurately detect trigger "受了伤"(is injured) if we can predict the entire nugget at character "受" or "伤". This redundant prediction makes our model more error-tolerant to character-wise classification errors, as verified in Section 4.4.

### 3.4 Model Learning

To train the trigger nugget generator, we regard all characters included in trigger nuggets as positive training instances, and randomly sample characters not in any trigger as negative instances and label them as "NIL". Suppose we have $T^G$ training examples in $S^G = \{(x_k, y_k^G)|k = 1, 2, ...T^G\}$ to train the trigger nugget generator, as well as $T^C$ examples in $S^C = \{(x_k, y_k^C)|k = 1, 2, ...T^C\}$ to train the event type classifier, we can define the loss function $\mathcal{L}(\theta)$ as follow:

$$\mathcal{L}(\theta) = - \sum_{(x_k, y_k^G) \in S^G} \log P(y_k^G|x_k;\theta) \\ - \sum_{(x_k, y_k^C) \in S^C} \log P(y_k^C|x_k;\theta) \quad (14)$$

where $\theta$ is parameters in NPNs. Since all modules in NPNs are differentiable, any gradient-based algorithms can be applied to minimize $\mathcal{L}(\theta)$.

## 4 Experiments

### 4.1 Data Preparation and Evaluation

We conducted experiments on two standard datasets: ACE2005 and TAC KBP 2017 Even-

| Model | ACE2005 | | | | | | KBPEval2017 | | | | | |
|---|---|---|---|---|---|---|---|---|---|---|---|---|
| | Trigger Identification | | | Trigger Classification | | | Trigger Identification | | | Trigger Classification | | |
| | P | R | F1 | P | R | F1 | P | R | F1 | P | R | F1 |
| FBRNN(Char) | 61.3 | 45.6 | 52.3 | 57.5 | 42.8 | 49.1 | 57.97 | 36.92 | 45.11 | 51.71 | 32.94 | 40.24 |
| DMCNN(Char) | 60.1 | 61.6 | 60.9 | 57.1 | 58.5 | 57.8 | 53.67 | 49.92 | 51.73 | 50.03 | 46.53 | 48.22 |
| C-BiLSTM* | 65.6 | 66.7 | 66.1 | 60.0 | 60.9 | 60.4 | - | - | - | - | - | - |
| FBRNN(Word) | 64.1 | 63.7 | 63.9 | 59.9 | 59.6 | 59.7 | 65.10 | 46.86 | 54.50 | 60.05 | 43.22 | 50.27 |
| DMCNN(Word) | 66.6 | 63.6 | 65.1 | 61.6 | 58.8 | 60.2 | 60.43 | 51.64 | 55.69 | 54.81 | 46.84 | 50.51 |
| HNN* | 74.2 | 63.1 | 68.2 | **77.1** | 53.1 | 63.0 | - | - | - | - | - | - |
| Rich-C* | 62.2 | 71.9 | 66.7 | 58.9 | 68.1 | 63.2 | - | - | - | - | - | - |
| KBP2017 Best* | - | - | - | - | - | - | 67.76 | 45.92 | 54.74 | **62.69** | 42.48 | 50.64 |
| NPN(Concat) | **76.5** | 59.8 | 67.1 | 72.8 | 56.9 | 63.9 | 64.58 | 50.31 | 56.56 | 59.14 | 46.07 | 51.80 |
| NPN(General) | 71.5 | 63.2 | 67.1 | 67.3 | 59.6 | 63.2 | 63.67 | 51.32 | 56.83 | 57.78 | 46.58 | 51.57 |
| NPN(Task-specific) | 64.8 | **73.8** | **69.0** | 60.9 | **69.3** | **64.8** | 64.32 | **53.16** | **58.21** | 57.63 | **47.63** | **52.15** |

Table 2: Experiment results on ACE2005 and KBPEval2017. * indicates the result adapted from the original paper. For KBPEval2017, "Trigger Identification" corresponds to the "Span" metric and "Trigger Classification" corresponds to the "Type" metric reported in official evaluation.

t Nugget Detection Evaluation (KBPEval2017) datasets. For ACE2005 (LDC2006T06), we used the same setup as Chen and Ji (2009), Feng et al. (2016) and Zeng et al. (2016), in which 569/64/64 documents are used as training/development/test set. For KBPEval2017, we evaluated our model on the 2017 Chinese evaluation dataset(LDC2017E55), using previous RichERE annotated Chinese datasets (LDC2015E78, LDC2015E105, LDC2015E112, and LDC2017E02) as the training set except 20 randomly sampled documents reserved as development set. Finally, there were 506/20/167 documents for training/development/test set. We used Stanford CoreNLP toolkit (Manning et al., 2014) to preprocess all documents for sentence splitting and word segmentation. Adadelta update rule (Zeiler, 2012) is applied for optimization.

Models are evaluated by micro-averaged Precision(P), Recall(R) and F1-score. For ACE2005, we followed Chen and Ji (2009) to compute the above measures. For KBPEval2017, we used the official evaluation toolkit [2] to obtain these metrics.

### 4.2 Baselines

Three groups of baselines were compared:

**Character-based NN models.** This group of methods solve Chinese Event Detection in a character-level sequential labeling paradigm, which include Convolutional Bi-LSTM model (C-BiLSTM) proposed by Zeng et al. (2016), Forward-backward Recurrent Neural Networks (FBRNN) by Ghaeini et al. (2016), and a character-level DMCNN model with a classifier using IOB encoding (Sang and Veenstra, 1999).

**Word-based NN models.** This group of methods directly adopt currently NN models into word-level sequences, which includes word-based FBRNN, word-based DMCNN and Hybrid Neural Network proposed by Feng et al. (2016), which incorporates CNN with Bi-LSTM and achieves the SOTA NN based result on ACE2005. To alleviate OOV problem stemming from word-trigger mismatch, we also adopt errata table replacing (Han et al., 2017), which introduce an errata table extracted from the training data and replace those words that part of whom was a trigger nugget with that trigger directly.

**Feature-enriched Methods.** This group of methods includes Rich-C (Chen and Ng, 2012) and CLUZH (KBP2017 Best) (Makarov and Clematide, 2017). Rich-C developed several handcraft Chinese-specific features, which is one of the state-of-the-art on ACE2005. CLUZH incorporated many heuristic features into LSTM encoder, which achieved the best performance in TAC KBP2017 evaluation.

### 4.3 Overall Results

Table 2 shows the results on ACE2005 and KBPEval2017. From this table, we can see that:

1) **NPNs steadily outperform all baselines significantly.** Compared with baselines, NPN(Task-specific) gains at least 1.6 (2.5%) and 1.5 (3.0%) F1-score improvements on trigger classification task on ACE2005 and KBPEval2017 respectively.

---
[2] github.com/hunterhector/EvmEval/tarball/master

2) **By exploiting compositional structures of triggers, our trigger nugget generator can effectively resolve the word-trigger mismatch problem.** As shown in Table 2, NPN(Task-specific) achieved significant F1-score improvements on trigger identification task on both datasets. It is notable that our method achieved a remarkable high recall on both datasets, which indicates that NPNs do detect a number of triggers which previous methods can not identify.

3) **By summarizing information from both characters and words, the hybrid representation learning is effective for event detection.** Comparing with corresponding character-based methods[3], word-based methods achieved 2 to 3 F1-score improvements, which indicates that words can provide additional information for event detection. By combining character-level and word-level features, NPNs are able to perform character-based event detection meanwhile take word-level knowledge into consideration too.

### 4.4 Comparing with Conventional Character-based Methods

To further investigate the effects of the trigger nugget generator, we compared NPNs with other character-based methods and analyzed behaviors of them. We conducted a supplementary experiment by replacing our trigger nugget generator and event type classifier with an IOB encoding labeling layer. We call this system NPN(IOB). Besides, we also compared the result with F-BRNN(Char), which proposes candidate trigger nuggets according to an external trigger table.

| Model | P | R | F1 |
|---|---|---|---|
| FBRNN(Char) | 57.97 | 36.92 | 45.11 |
| NPN(IOB) | 60.96 | 47.39 | 53.32 |
| NPN(Task-specific) | **64.32** | **53.16** | **58.21** |

Table 3: Performances of character-based methods on KBP2017Eval Trigger Identification task.

Table 3 shows the results on KBP2017Eval. We can see that NPN(Task-specific) outperforms other methods significantly. We believe this is because:

1) FBRNN(Char) only regards tokens in the candidate table as potential trigger nuggets, which

---

[3]C-BiLSTM and HNN are similar methods to some extent. They both use a hybrid representation from CNN and BiLSTM encoders.

---

limits the choice of possible trigger nuggets and results in a very low recall rate.

2) To accurately identify a trigger, NPN(IOB) and conventional character-based methods require all characters in a trigger being classified correctly, which is very challenging (Zeng et al., 2016): many characters appear in a trigger nugget will not serve as a part of a trigger nugget in the majority of contexts, thus they will be easily classified into "NIL". For the first example in Table 5, NPN(IOB) was unable to fully recognize the trigger nugget "贺电"(congratulatory message) because character "贺"(congratulatory) doesn't often serve as part of "PhoneWrite" trigger. In fact, "贺" serves as a "NIL" in the majority of similar contexts, e.g., "贺喜"(congratulation) and "祝贺"(congratulation).

3) NPNs are able to handle above problems. First, NPNs doesn't rely on candidate tables to generate potential triggers, which guarantees a good generalization ability. Second, NPNs propose the entire trigger nugget at each character, such a redundant prediction paradigm makes NPNs more error-tolerant to character-level errors. For example, even might mistakenly classify "贺" into "NIL", NPNs can still identify the correct nugget "贺电" at character "电" because "电" is a common part of "PhoneWrite" event trigger.

### 4.5 Influence of Word-Trigger Mismatch

This subsection investigates the effects of resolving the word-trigger mismatch problem using different methods. According to different types of word-trigger match, we split KBP2017Eval test set into three parts: Exact, Part-of-Word, Cross-Words, which are as defined in Table 1.

| Model | Exact | Part | Cross |
|---|---|---|---|
| NPN(IOB) | 48.65 | 29.13 | 8.54 |
| DMCNN(Word) | 57.36 | 23.28 | 0.00 |
| - w/o Errata replacing | 59.03 | 0.00 | 0.00 |
| NPN(Task-specific) | 56.47 | 42.66 | 26.58 |

Table 4: Recall rates on three word-trigger match splits on KBP2017Eval Trigger Identification task.

Table 4 shows the recall of different methods on each split. NPN(Task-specific) significantly outperform other baselines when trigger-word mismatch exists. This verified that NPNs can resolve different cases of word-trigger mismatch problems robustly, meanwhile retain high performance on exact match cases. In contrast, NPN(IOB) can not exactly detect boundaries of trigger nuggets, thus has a low recall on all splits. Conventional DMCNN regards words as potential triggers, which means it can only identify triggers that exactly match with words. As the second example in Table 5, word "死伤"(dead or injured) as a whole has never been annotated as a trigger, so DMCNN is unable to recognize it at all. Errata replacing can only solve some of the part-of-word mismatch problem, but it can not handle the cases where one word contains multiple triggers(e.g., "死伤" in Table 5) and the cases that a trigger crosses multiple words.

| Sentence | DMCNN | NPN(IOB) | NPN | Correct |
|---|---|---|---|---|
| 贺电/全文/如下,... Full **congratulatory message**:... | (贺电,PhoneWrite) | (电,PhoneWrite) | (贺电,PhoneWrite) | (贺电,PhoneWrite) |
| 死伤/的/所有/士兵... all soldiers **died and injured**... | None | (死,Die) (伤,Injure) | (死,Die) (伤,Injure) | (死,Die) (伤,Injure) |

Table 5: System prediction examples. (X,Y) indicates a trigger nugget X is annotated with event type Y.

### 4.6 Effects of Hybrid Representation

This section analyzed the effect of feature hybrid in NPNs. First, from Table 2, we can see that Task-specific Hybrid method achieved the best performance in both datasets. Surprisingly, simple Concat Hybrid outperforms the General Hybrid approach. We believe this is because the trigger nugget generator and the event type classifier rely on different information, and therefore using one unified gate is not enough. And Task-specific Hybrid uses two different task-specific gates which can satisfy both sides, thus resulting in the best overall performance.

Furthermore, to investigate the necessary of using hybrid features, an auxiliary experiment, called NPN(Char), was conducted by removing word-level features from NPNs. Also, we compared with the model removing character-level features, which is the original DMCNN(Word).

| Model | P | R | F1 |
|---|---|---|---|
| DMCNN(Word) | 54.81 | 46.84 | 50.51 |
| NPN(Char) | 56.19 | 43.88 | 49.28 |
| NPN(Task-specific) | **57.63** | **47.63** | **52.15** |

Table 6: Results of using different representation on Trigger Classification task on KBP2017Eval.

Table 6 shows the experiment results. We can see that neither character-level or word-level representation can achieve competitive results with the NPNs. This verified the necessity of hybrid representation. Besides, we can see that NPN(Char) outperforms other character-level methods in Table 2, which further confirms that our trigger nugget generator is still effective even only using character-level information.

## 5 Related Work

Event detection is an important task in information extraction and has attracted many attentions. Traditional methods (Ji and Grishman, 2008; Patwardhan and Riloff, 2009; Liao et al., 2010; McClosky et al., 2011; Hong et al., 2011; Huang and Riloff, 2012; Li et al., 2013a,b, 2014) rely heavily on hand-craft features, which are hard to transfer among languages and annotation standards.

Recently, deep learning methods, which automatically extract high-level features and perform token-level classification with neural networks (Chen et al., 2015b; Nguyen and Grishman, 2015), have achieved significant progress. Some improvements have been made by jointly predicting triggers and arguments (Nguyen et al., 2016) and introducing more complicated architectures to capture larger scale of contexts (Feng et al., 2016; Nguyen and Grishman, 2016; Ghaeini et al., 2016). These methods have achieved promising results in English event detection.

Unfortunately, the word-trigger mismatch problem significantly undermines the performance of word-level models in Chinese event detection (Chen and Ji, 2009). To resolve this problem, Chen and Ji (2009) proposed a feature-driven BIO tagging methods at character-level sequences. Qin et al. (2010) introduced a method which can automatically expand candidate Chinese trigger set. While Li et al. (2012) and Li and Zhou (2012) defined manually character compositional patterns for Chinese event triggers. However, their methods rely on hand-crafted features and patterns, which make them difficult to be integrated into recent Deep Learning models.

Recent advances have shown that neural networks can effectively capture spatial and positional information from raw inputs (Ren et al., 2015; He et al., 2017; Wang and Jiang, 2017).

This paper designs Nugget Proposal Networks to capture character compositional structure of event triggers, which is more robust and more effective than previous hand-crafted patterns or character-level sequential labeling methods.

## 6 Conclusions and Future Work

This paper proposes Nugget Proposal Networks for Chinese event detection, which can effectively resolve the word-trigger mismatch problem by modeling and exploiting character compositional structure of Chinese event triggers, using hybrid representation which can summarize information from both characters and words. Experiment results have shown that our method significantly outperforms conventional methods.

Because the mismatch between words and extraction units is a common problem in information extraction, we believe our method can also be applied to many other languages and tasks for exploiting inner composition structure during extraction, such as Named Entity Recognition.

## Acknowledgments

This work is supported by the National Natural Science Foundation of China under Grants no. 61433015, 61572477 and 61772505, and the Young Elite Scientists Sponsorship Program no. YESS20160177. Moreover, we sincerely thank all reviewers for their valuable comments.

## References

Chen Chen and Vincent Ng. 2012. Joint modeling for chinese event extraction with rich linguistic features. In *Proceedings of COLING 2012*.

Xinxiong Chen, Lei Xu, Zhiyuan Liu, Maosong Sun, and Huanbo Luan. 2015a. Joint learning of character and word embeddings. In *Proceedings of IJCAI 2015*.

Yubo Chen, Liheng Xu, Kang Liu, Daojian Zeng, and Jun Zhao. 2015b. Event extraction via dynamic multi-pooling convolutional neural networks. In *Proceedings of ACL 2015*.

Zheng Chen and Heng Ji. 2009. Language specific issue and feature exploration in chinese event extraction. In *Proceedings of NAACL-HLT 2009*.

Xiaocheng Feng, Lifu Huang, Duyu Tang, Bing Qin, Heng Ji, and Ting Liu. 2016. A language-independent neural network for event detection. In *Proceedings of ACL 2016*.

Reza Ghaeini, Xiaoli Z Fern, Liang Huang, and Prasad Tadepalli. 2016. Event nugget detection with forward-backward recurrent neural networks. In *Proceedings of ACL 2016*.

Xianpei Han, Xiliang Song, Hongyu Lin, Qichen Zhu, Yaojie Lu, Le Sun, Jingfang Xu, Mingrong Liu, Ranxu Su, Sheng Shang, Chenwei Ran, and Feifei Xu. 2017. ISCAS_Sogou at TAC-KBP 2017. In *Proceedings of TAC 2017*.

Kaiming He, Georgia Gkioxari, Piotr Dollár, and Ross Girshick. 2017. Mask r-cnn. *arXiv preprint arXiv:1703.06870*.

Yu Hong, Jianfeng Zhang, Bin Ma, Jianmin Yao, Guodong Zhou, and Qiaoming Zhu. 2011. Using cross-entity inference to improve event extraction. In *Proceedings of ACL-HLT 2011*.

Ruihong Huang and Ellen Riloff. 2012. Modeling textual cohesion for event extraction. In *Proceedings of AAAI 2012*.

Heng Ji and Ralph Grishman. 2008. Refining event extraction through cross-document inference. In *Proceedings of ACL 2008*.

Peifeng Li and Guodong Zhou. 2012. Employing morphological structures and sememes for chinese event extraction. In *Proceedings of COLING 2012*.

Peifeng Li, Guodong Zhou, Qiaoming Zhu, and Libin Hou. 2012. Employing compositional semantics and discourse consistency in chinese event extraction. In *Proceedings of EMNLP-CoNLL 2012*.

Peifeng Li, Qiaoming Zhu, and Guodong Zhou. 2013a. Argument inference from relevant event mentions in chinese argument extraction. In *Proceedings of ACL 2013*.

Qi Li, Heng Ji, Yu HONG, and Sujian Li. 2014. Constructing information networks using one single model. In *Proceedings of EMNLP 2014*.

Qi Li, Heng Ji, and Liang Huang. 2013b. Joint event extraction via structured prediction with global features. In *Proceedings of ACL 2013*.

Shasha Liao, New York, Ralph Grishman, and New York. 2010. Using document level cross-event inference to improve event extraction. In *Proceedings of ACL 2010*.

Peter Makarov and Simon Clematide. 2017. UZH at TAC KBP 2017: Event nugget detection via joint learning with softmax-margin objective. In *Proceedings of TAC 2017*.

Christopher D. Manning, Mihai Surdeanu, John Bauer, Jenny Finkel, Steven J. Bethard, and David McClosky. 2014. The Stanford CoreNLP natural language processing toolkit. In *Proceedings of ACL 2014*.

David McClosky, Mihai Surdeanu, and Christopher D Manning. 2011. Event extraction as dependency parsing. In *Proceedings of ACL-HLT 2011*.

Thien Huu Nguyen, Kyunghyun Cho, and Ralph Grishman. 2016. Joint event extraction via recurrent neural networks. In *Proceedings of NAACL-HLT 2016*.

Thien Huu Nguyen and Ralph Grishman. 2015. Event detection and domain adaptation with convolutional neural networks. In *Proceedings of ACL 2015*.

Thien Huu Nguyen and Ralph Grishman. 2016. Modeling skip-grams for event detection with convolutional neural networks. In *Proceedings of EMNLP 2016*.

Siddharth Patwardhan and Ellen Riloff. 2009. A unified model of phrasal and sentential evidence for information extraction. In *Proceedings of EMNLP 2009*.

Bing Qin, Yanyan Zhao, Xiao Ding, Ting Liu, and Guofu Zhai. 2010. Event type recognition based on trigger expansion. *Tsinghua Science and Technology*, 15(3):251–258.

Shaoqing Ren, Kaiming He, Ross Girshick, and Jian Sun. 2015. Faster r-cnn: Towards real-time object detection with region proposal networks. In *Proceedings of NIPS 2015*.

Erik F. Tjong Kim Sang and Jorn Veenstra. 1999. Representing text chunks. In *Proceedings of EACL 1999*.

Shuohang Wang and Jing Jiang. 2017. Machine comprehension using match-lstm and answer pointer. In *Proceedings of ICLR 2017*.

Matthew D. Zeiler. 2012. Adadelta: An adaptive learning rate method. *arXiv preprint arXiv:1212.5701*.

Ying Zeng, Honghui Yang, Yansong Feng, Zheng Wang, and Dongyan Zhao. 2016. A convolution bilstm neural network model for chinese event extraction. In *Proceedings of NLPCC-ICCPOL 2016*.